\documentclass[conference]{IEEEtran}
\IEEEoverridecommandlockouts
\usepackage{cite}
\usepackage{float}
\usepackage{amsmath,amssymb,amsfonts}
\usepackage{adjustbox}
\usepackage{adjustbox}
\usepackage{times}
\usepackage{verbatim}
\usepackage{enumitem}
\usepackage{graphicx}
\usepackage{textcomp}
\usepackage{xcolor}
\usepackage{todonotes}
\usepackage[english]{babel}
\usepackage{blindtext}
\usepackage{subfigure}
\usepackage{graphicx}

\def\BibTeX{{\rm B\kern-.05em{\sc i\kern-.025em b}\kern-.08em
    T\kern-.1667em\lower.7ex\hbox{E}\kern-.125emX}}
\begin{document}

\title{Cyber-Physical Testbed for Human-Robot Collaborative Task Planning and Execution }

\author{\IEEEauthorblockN{ Tuly Hazbar$^{1}$ \qquad Shitij Kumar$^{2}$\qquad Ferat Sahin$^{3}$}
\\
\IEEEauthorblockA{Department of Electrical Engineering \\
Rochester Institute of Technology\\
Rochester, NY, 14623, USA \\
tmh6831$^{1}$,spk4422$^{2}$,feseee$^{3}$@rit.edu }
}

\maketitle

\begin{abstract}In this paper, we present a cyber-physical testbed created to enable a human-robot team to perform a shared task in a shared workspace. The testbed is suitable for the implementation of a tabletop manipulation task, a common human-robot collaboration scenario. The testbed integrates elements that exist in the physical and virtual world. In this work, we report the insights we gathered throughout our exploration in understanding and implementing task planning and execution for human-robot team. 
\end{abstract}


\section{Introduction}
Robots that can collaborate with humans present a clear added value to the industry. They promise to be a solution for performing tasks that are hard to fully automate and for industries that undergo frequent changes in their production line, where reconfigurability and adaptability are of great importance \cite{Michalos2014}. By teaming with a human, the collaboration benefits from the decision-making capabilities of the human in selecting the appropriate actions to be executed for a given task, consequently increase the overall flexibility and adaptability of the human-robot team. However, counting on the human's decision-making skills by having the human plan for his/her own actions and the robot's actions does not make the robot a collaborator but a recipient of human commands. This type of interaction becomes a turn taking based, which has many disadvantages towards the overall performance of the team; one apparent drawback is the decrease in productivity of the team. 
\begin{figure}[t!]
    \centering
	\includegraphics[width=0.5\textwidth]{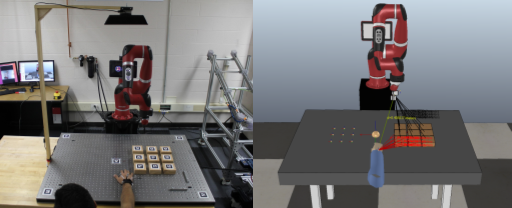}
	\caption{The physical and virtual representation of the HRC scenario. The human-robot team is moving the workpieces from one side of the table to the other while maintaining the same arrangement.}
	\label{fig:digital_twin}
\end{figure}

Although collaborative robots have become safe and reliable enough to operate close to humans, human-robot teaming still lacks many aspects that make the collaboration successful, especially when compared to a human-human team executing a shared task. Typically, when a group of two humans works together, they display a high level of action selection and coordination, and on the fly work distribution and scheduling. One aim of the HRC research is to reach this level of coordination referred to as fluency in human-robot teams \cite{Hoffman2013a}.  In such a team the two partners are capable of planning their own actions, and coordinate their execution with one another.  Similar to other research in HRC, we believe that the transition from robots as recipients of human instructions to robots as capable collaborators hinges around their ability to select their own action and coordinate its execution with a human partner\cite{Baraglia2017}. The HRC scenario this paper studies, is where the human and the robot share a workspace and the robot is expected to achieve multiple object manipulations by taking into account, at every stage, the actions of the human partner (see Figure \ref{fig:digital_twin}). The robot must be able to move and act in a safe, efficient, and fluent way. 

 The human is far superior to its robot counterpart in terms of both perception and dexterity. It is important to compensate for the unbalance of capabilities as a first step towards studying and improving any aspect in any HRC scenario. For this purpose, we need to have a setup that supports studying and implementing the strategies we envision for improving task planning and execution for human-robot team. The contributions of our research are:
 1) Outline the main elements of an HRC testbed that enables a human and a robot to perform a shared task in a shared workspace; 
 2) recommendations for the design of such testbeds;
 3) reporting our insights with a case study.
 
The remainder of the paper is organized as follows: Section \ref{sec:background} provides relevant background information for this research. Section \ref{sec:approach} describes the proposed approach for creating a cyber-physical testbed for human-robot collaboration. Implementation of the testbed is explained in \ref{sec:Implementation} and a case study is presented in Section \ref{sec:casestudy}. Conclusions are future work in Section \ref{sec:Conclusions}.

\section{Background and Related Work}
\label{sec:background}
 This section provides background information and literature review of related research in task planning and execution for a human-robot team. It summarizes some of the existing practices for human-robot collaboration setups in research and industry. A brief discussion of similar and related research of HRC platforms is also presented.

\subsection{Task Planning for Human-Robot Collaboration}
Task planning is a key ability for intelligent robotic systems, increasing their autonomy through the construction of sequences of actions to achieve a final goal. However, when working in a team, planning for the sequence of actions without taking into consideration the actions of the other partner is not enough, especially when collaborating in achieving a common task. Therefore, adaptability and flexibility of planning are crucial in such scenarios. This has been investigated extensively in previous work. Our testbed is inspired by this work \cite{Lemaignan2017a}, that offers a comprehensive system that identifies and integrates individual and collaborative human inspired cognitive skills a robot should have to share space and task with a human partner. The scenario we are interested in is similar to the one investigated by \cite{Pellegrinelli2016c}, where a robot and a human manipulate an overlapping set of objects, without using the same object simultaneously. A key element in task planning and execution is the robot ability to recognized and anticipate the human's actions. Similar to other work in the literature \cite{PrezDArpino2015FastTP}\cite{Zanchettin2017b}, we recognize and anticipate human actions through the human arm motion within the workspace to enable task planning and execution.

\subsection{HRC Testbed Requirements }

Previous work in the literature emphasizes the need for a testbed that enables implementing and studying different aspects of human-robot collaboration\cite{kirsch10testbed}\cite{Zeylikman2017}. There are standard practices followed within the HRC research to create setups for HRC scenarios — having multiple layers inspired by the field of cognitive robotics, such as human-level perception and human-like decision-making ability. Testing and implementing robot behaviors in a simulated environment is also common since it provides a controlled environment that helps in studying and exploring one aspect of HRC in isolation\cite{Darvish2018}. What we don't see extensively used in the HRC setups, is a complete virtual model of an entire HRC scenario the exists in the physical world. This concept is referred to as the Digital Twin in industry. A digital twin is a comprehensive physical and functional description of a component, product, or system virtually. It has been used for surveillance, evaluation, planning, and manipulation of the production environment \cite{Boschert2016}. The work in \cite{Cichon2017} is an example where a digital twin is utilized in human-robot context. We believe combining physical and virtual worlds can lead to a new approach on how to implement HRC scenarios. In Section \ref{sec:approach}, we will explain the added value of the virtual replica of the physical world as an element of an HRC testbed. 
\begin{figure}[h!]
    \centering
    \includegraphics[width=8.5cm,keepaspectratio]{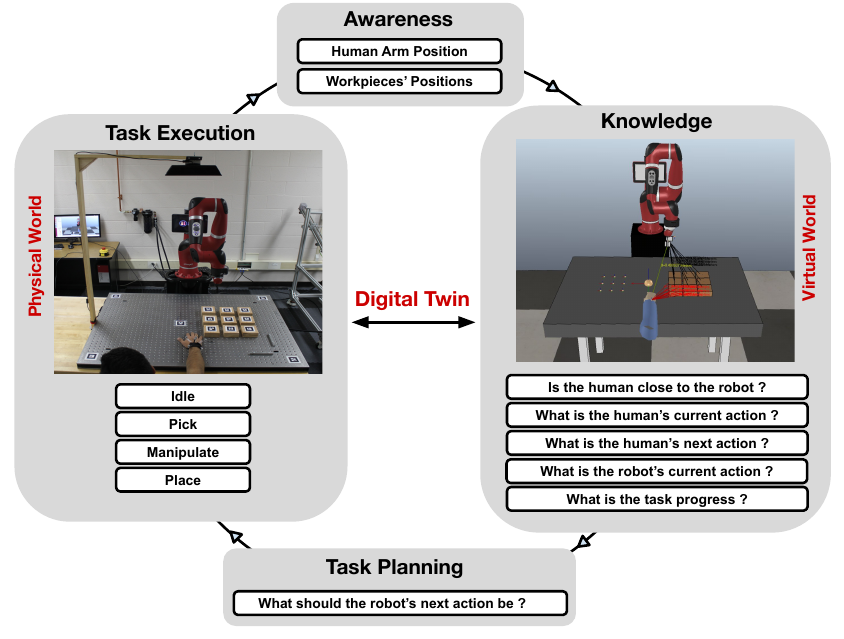}
    \caption{Overview of the HRC cyber-physical testbed. It is composed of 4 main modules. These modules are integrated through the digital twin concept.}
    \label{fig:overview}
\end{figure}

\section{Cyber-Physical HRC Testbed}
\label{sec:approach}
This section explains our approach to create a testbed that supports human-robot collaborative task planning and execution. It outlines the main elements that make a testbed for that purpose.

\subsection{Perception}

For the robot to be a better collaborator, it should be aware of the human partner and his/her actions and the progress of the shared task. Using sensors, the robot will have a perception of these two components, and through perception, the knowledge is defined in the next section. 

\subsubsection{Perception of the human arm trajectory}
 Human arm position tracking is crucial for our HRC scenario. This information will be utilized by other parts of the testbed to calculate and define the following: (a) human-workpieces euclidean distances to infer the human goal (b) human current arm position for safety purposes.  
\subsubsection{Perception of the workpieces} 
For tabletop manipulation task, it is essential to know the position of the workpieces in the workspace at all times. The robot uses this information for motion planning and defining task progress to select its action.
\subsection{Forming Knowledge Through Digital Twin}
In general, the fundamental elements that characterize an HRC scenario include the following: the agents, the work environment, and the workpieces. Every element in the physical world that affects the HRC scenario has a counterpart in the virtual world that mirrors its state in real-time. As seen in figure \ref{fig:overview}, the input to the virtual world is the data acquired from the perception component (robot joint states, human arm position, and workpieces positions), and the output of the virtual world is the integrated knowledge concerning the human actions, the robot actions, and the task progress. We believe this explicit integration of all components in a virtual platform is useful for the following reasons: 
\begin{list}{\textbullet}{\leftmargin=0.5em}

\item Forming comprehensive knowledge about all the components that make up the HRC scenario through the use of the physics engines, and the calculation modules available in the virtual platform. This facilitates relating both the human and the robot activity with each other in real time. Also, depicting the agents' interaction with the workpieces and the overall task progress. This type of knowledge is vital for robot autonomy, enabling collaboration, and enhancing its quality.   

\item A virtual representation of the workspace can be utilized to set spatial limits in the virtual world that are necessary for the robot during the interaction. Also, define the description of the end goal of the shared task in the virtual world that is necessary for the robot to complete the task — for example, markers on the target locations of where the workpieces need to be placed. This is used by the robot for path planning. We can eliminate the need for adding markers in the physical world, which is specifically harder to control in an industrial setting compared to a lab environment.

\item Real-time HRC diagnostic through visualizing and monitoring the state of the human, the robot, the workspace, and the workpieces. This can serve as a way to aggregate and organize the data acquired in a way that can support researchers in the process of analyzing the robot behavior during and post the interaction. 

\item Although the virtual world does not account for the uncertainties of the real world, virtual testing is proven to be a powerful approach for testing HRC algorithms being developed. Having a realistic data of the human arm trajectory and actions when performing the task can be used to compare different robot behaviors against it virtually. This gives valuable insight into the robot behavior being developed through in-depth analysis while maintaining human safety.

\item Metrics that are necessary for evaluating the HRC scenario can be easily and accurately found in the virtual world (see Section \ref{sec:casestudy}). It is an alternative to some common practices used in the HRC research that are subjected to human error during observing the collaboration or analyzing it after it took place through a video recording. The metrics are found right after the task is completed.
\end{list}

For the reasons mentioned above, we believe the digital twin is a vital factor in supporting the design, build, control, monitor, and evaluation of an HRC scenario. We consider it one of the main elements in our HRC testbed.
\subsection{ Robot Intelligence }
The robot needs to plan and execute -on the fly- an action from a set of possible actions taking into consideration the human activity. We will explain the three main types of decisions the robot can make: immediate decisions concerning human safety, proactive and reactive decisions regarding collaboration fluency and the task final goal. Then we will explain the formalism used to implement this behavior. The distinction between task planning and execution is blurred since planning, and execution occur intermixed at various levels, but we have them separated below for clarification purposes.

\subsubsection{ Robot Reasoning for Task Planning }  
With the use of the acquired knowledge, the robot can decide which workpiece it should manipulate and select an action to perform accordingly. The selection process will be in favor of minimizing the disruption of human activity when sharing the workspace and workpieces to be manipulated. Therefore, the robot should have the ability to reason about which workpiece the human is going to choose to manipulate, refer to as the human goal. Then the robot should eliminate the human goal from the set of workpieces it can manipulate, and choose the closest workpiece relative to its current end effector position. Recognizing the human goal is vital in the robot goal selection process. In our HRC scenario, it is based on the Euclidean distance between the human arm position against the workpieces on the workspace. A probabilistic approach can be used to assign probabilities on the workpieces that need to be manipulated; the robot can utilize that in its goal selection process. The robot should also reason about the safety of the human partner at all times. In our current work, the robot selects to be idle when the human-robot minimum distance is below a certain threshold. 

\subsubsection{ Robot Task Execution}
After goal selection, the robot starts executing a sequence of actions needed to place the workpiece in the location defined by the task's end goal. During execution, the robot should still be aware of the human partner actions and arm position within the workspace to adapt to changes that require the robot to change its action on the fly. To execute this behavior, we use a formalism that combines Concurrency with Hierarchical State Machines, to account for situations where interrupting the current action is necessary to respond to something more important.

\section{Implementation }
\label{sec:Implementation}
We implemented our system on a Sawyer, Rethink Robotics research robot, using the Robot Operating System - ROS \cite{Quigley}. Suited for our HRC scenario, the robot can perform basic actions like pick and place of workpieces. We have utilized Moveit - a ROS package for the robot motion planning. The perception of the workpieces is based on ArUco library, an open source library to generate and detect fiducial markers. Each workpiece in the workspace is equipped with a unique ID associated with a specific marker. To cover the entire workspace with minimum occlusion of the workpieces, we mounted a Kinect on top of the workspace as shown in figure \ref{fig:digital_twin}. The human arm position is tracked using the OptiTrack 3D tracking system. OptiTrack tracks the human hand using cameras with low-latency and high precision \cite{optitrack}. This system requires the human partner to have an on body markers; therefore the human partner wears a wrist band shown in Figure \ref{fig:digital_twin} on the hand he/she prefers to use while performing the task. We choose the Virtual Robot Experimentation Platform (V-REP) to create our virtual HRC scenario \cite{rohmerVREPVersatileScalable2013}.

\begin{figure}[ht]
	\centering
	\includegraphics[width=0.5\textwidth]{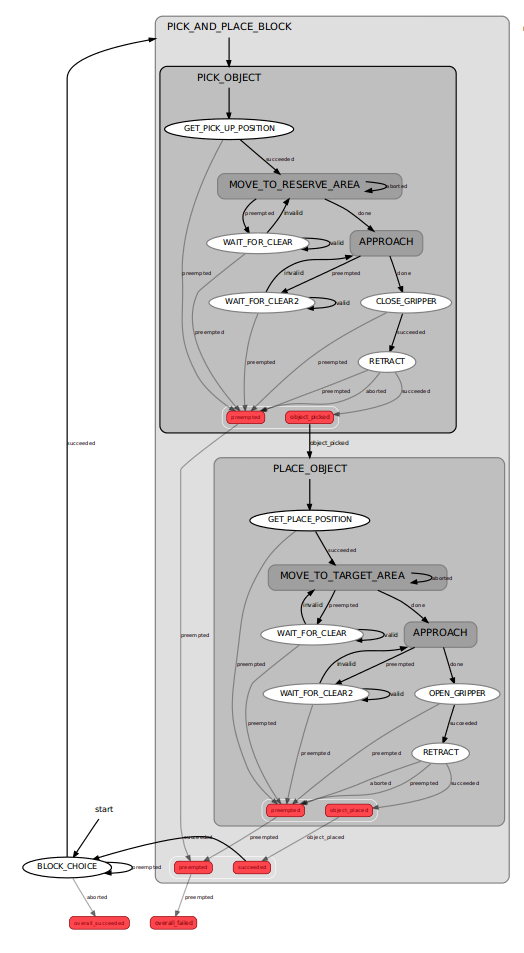}
	\caption{The plan execution that combines concurrency with hierarchical state machines for Pick and Place Tasks.}
		\label{fig:HFSM}
\end{figure}

We use SMACH for robot task execution. SMACH is a Python library that provides structures based on hierarchical concurrent state machines. Figure \ref{fig:HFSM}  shows the plan execution levels. The execution of the manipulating a workpiece starts when the BLOCK-CHOICE state receives a response from the node responsible for selecting the robot goal. Concurrency in the state machine takes places in the MOVE-TO-RESERVE-AREA, APPROACH, MOVE-TO-TARGET-AREA and APPROACH where two states execute simultaneously. One state preforms the manipulation action and the second state monitors the human-robot distance. The action state will be preempted when the distance is below a predefined threshold and resume when the condition does not hold anymore.

\section{Case Study}
\label{sec:casestudy}

A testbed that enables us to study and implement task planning and execution for a human-robot team is crucial. We are particularly interested in understating action selection and coordination on the fly for such a team that shares the workspace and the workpieces that need to be manipulated.

\subsection{Experiment Design}
There are nine workpieces that the human and the robot need to arrange in a specific order based on a given end goal. For this case study, the participant is given a model pattern of workpieces to move from one side of the table to the other within the workspace, maintaining the same arrangement. This HRC scenario does not put any restrictions on the participant on which workpieces he/she can manipulate, or create exclusive workspace zones for any of team members. This will help in creating a similar human-human team kind of interaction that we hypothesize is more fluent. 

\subsection{Results}

The following Figure \ref{fig:results} shows snippets of the collaboration using the cyber-physical testbed at different instances. The first row represents the start state, and the last represents the task final state. Instances of a human concurrently picking and placing a workpiece with the robot is portrayed in Figure \ref{fig:results}.
\begin{figure*}[h!]
    \centering
    \includegraphics[width=1\textwidth]{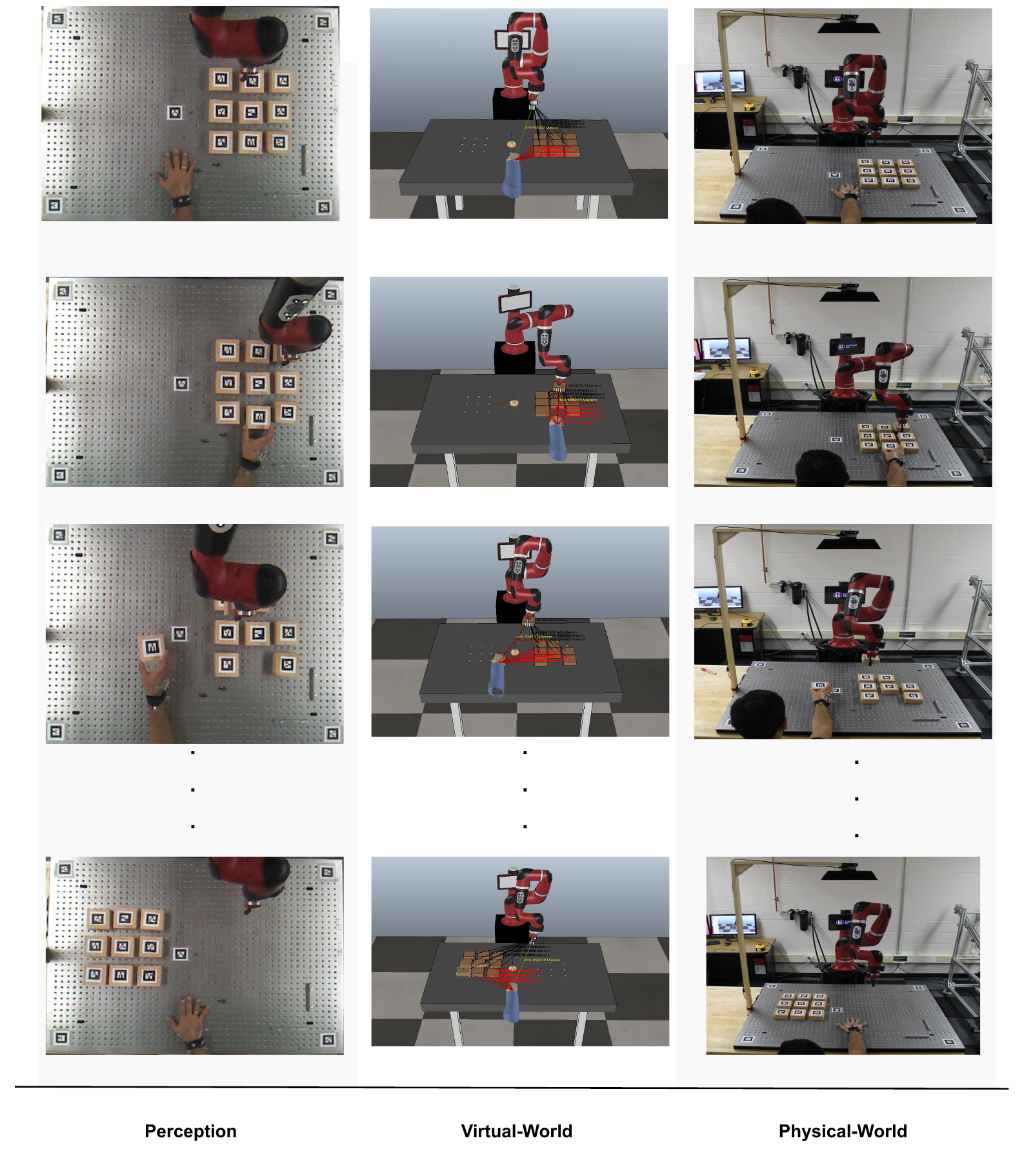}
    \caption{  The sequence of actions associated with the experiment of performing the collaborative task. The first column of the frames represents the Perception view from the camera of the shared workspace. The second column shows the virtual world representation i.e. digital twin of human, robot, and objects, as shown in the physical world (Last Column). The last row shows the completion of the human-robot task. }
    \label{fig:results}
\end{figure*}

\subsection{Metrics - Evaluation of Human-Robot Collaboration }
We choose to evaluate the collaboration through both objective and subjective metrics. The following explains the objective metrics of interest: 
\begin{itemize}
    \item \textbf{Robot Idle Time} Percentage of time out of the total task time, during which the robot has been not active. The robot can be idle due to predefined rules to prevent the human-robot collision. 
    \item \textbf{Human Idle Time} Percentage of time out of the total task time, during which the human has been not active.
    \item \textbf{Number of Collisions} between human and robot :(\textit{Note: Collision is defined  as the event where the minimum distance between human and robot is below a predefined threshold.})  
    \item \textbf{Functional Delay:} Percentage of time out of the total task time, between the end of one agent's action and the beginning of the other agent's action.
    \item \textbf{Concurrent activity:} Percentage of time out of the total task time, during which both agents have been active at the same time.
    \item \textbf{Number of actions performed by each agent}
    \item \textbf{Time to complete task}: i.e. the time taken to place the 9 workpieces in their target locations.
\end{itemize}
These metrics are based on the guidelines set by \cite{Hoffman2013a} to evaluate the fluency of the human-robot collaboration and the insights of\cite{6281347} to improve human-robot team performance.  

\section{Conclusion}
\label{sec:Conclusions}
In this paper, a cyber-physical testbed was created to enable a human-robot team to perform a shared task in a collaborative workspace. A digital twin of the HRC scenario was created and used during a human-robot collaborative tabletop manipulation task in a shared workspace. This setup was successfully used to test robot task planning and execution based on the states represented and reported using the digital twin, thus validating the importance of using a virtual world representation of all actors: human, robots, and objects during a human-robot collaborative task and its significance in task execution and planning. 

Our ongoing work is performing human-subject experiments and evaluating them based on the proposed metrics. We wish to gather more insights through our explorations of understanding and implementing task planning and execution for human-robot team. 

\section*{Acknowledgment}
The authors would like to thank the staff of Multi Agent Bio-Robotics Laboratory (MABL) and the CM Collaborative Robotics Research (CMCR) Lab for their valuable inputs.

\bibliographystyle{IEEEtran}
\bibliography{smc2019_tulyHazbar}
\bibliographystyle{plain}

\end{document}